\documentclass{article}

\usepackage[preprint]{neurips_2024}

\usepackage[utf8]{inputenc} 
\usepackage[T1]{fontenc}    
\usepackage{hyperref}       
\usepackage{url}            
\usepackage{booktabs}       
\usepackage{amsfonts}       
\usepackage{nicefrac}       
\usepackage{microtype}      
\usepackage{xcolor}         
\usepackage{tabularx}
\usepackage{multirow}
\usepackage{subfigure}
\usepackage{caption}
\usepackage[pdftex]{graphicx}
\usepackage{adjustbox}
\usepackage{amsmath} 
\usepackage{amssymb}
\title{Dynamic Motion Synthesis: Masked Audio-Text Conditioned Spatio-Temporal Transformers}

\author{%
  Sohan Anisetty\\
  Department of Computer Science\\
  Georgia Tech\\
  \texttt{sanisetty3@gatech.edu} \\
  \And
  James Hays \\
  Department of Computer Science \\
  Georgia Tech \\
  \texttt{hays@gatech.edu} \\
}

\begin{document}

\maketitle

\begin{abstract}
Our research presents a novel motion generation framework designed to produce whole-body motion sequences conditioned on multiple modalities simultaneously, specifically text and audio inputs. Leveraging Vector Quantized Variational Autoencoders (VQVAEs) for motion discretization and a bidirectional Masked Language Modeling (MLM) strategy for efficient token prediction, our approach achieves improved processing efficiency and coherence in the generated motions. By integrating spatial attention mechanisms and a token critic we ensure consistency and naturalness in the generated motions. This framework expands the possibilities of motion generation, addressing the limitations of existing approaches and opening avenues for multimodal motion synthesis.

\end{abstract}

\section{Introduction}

The field of motion generation aimed at producing continuous, natural, and logical human movements based on control conditions has garnered considerable attention. Current research is divided into text based motion generation\cite{tomato,t2mgpt,momask,tm2t,motiongpt,mdm,mld,motiondiffuse}, music based dance generation\cite{dancing2music,motionet,charmotsyn,learning2dance_gcn,bailando,music2dance,genrecond,aichoreo,youneverstop,attenspatiotemporal,dancerevolution, tm2d,choreomaster}, and speech and transcript based gesture generation\cite{emage,gesturediffuclip}. Each modality presents unique challenges and has been approached with modality-specific methods, resulting in limited crossover between subdomains. However, this compartmentalization leaves numerous possibilities unexplored. For instance, consider the scenario of generating motion following speech audio and transcript while executing cartwheels—an oddly specific yet illustrative example highlighting the limitations of existing models, which fail to integrate diverse modalities like audio and text seamlessly.

The landscape of motion generation research is characterized by two main approaches: diffusion-based and language model-based methods. Diffusion-based models extend traditional image generation techniques but noise and denoise 1D motion sequences instead of 2D images. On the other hand, language model(LM)-based approaches convert motion into discrete tokens, treating them similarly to language tokens. Existing motion generation methods predominantly rely on standard auto-regressive transformers\cite{transformers} for token prediction in a unidirectional manner. While causal attention models intuitively capture the temporal nature of motion, they exacerbate tokenization errors by relying solely on previous tokens for prediction. Moreover, these models face computational challenges, particularly as motion and context length increases, and lack global context for tasks such as motion inpainting and editing.

Additionally, we extend prior research by modeling not only the body but also the hands. A naive approach that increases the codebook size would be in vain\cite{soundstream} and a more practical option would be increasing the number of codebooks. Multi codebook generation is relatively less explored. Recent methods flatten multiple codebooks into a single sequence, sacrificing spatial relations and increasing computation. While MusicGen\cite{musicgen} has experimented with multi codebook generation, they take advantage of the next token prediction scheme of auto regressive transformer\cite{transformers}, something not available to masked language modelling based models.  

Building upon these observations, we introduce a motion generation framework capable of producing motion sequences of arbitrary length, conditioned on multiple modalities simultaneously, and adaptable over time.

Our work is grounded in the assumption that motion is physically constrained and can be represented as a weighted combination of motion primitives. To this end, we employ Vector Quantized Variational Autoencoders (VQVAEs)\cite{vqvae} to discretize motion into tokens and leverage language models for conditional token prediction. This formulation offers several advantages, including the utilization of Large Language Model (LLM) research and optimizations, the capacity to encode multiple modalities in a common representation for improved performance in multi-modal reasoning tasks, and enhanced robustness and generalization through token-based model inputs\cite{discretevisionstrength}. 

To address the limitations of auto-regressive approaches, we utilize a bidirectional Masked Language Modeling (MLM) strategy, which has proven effective in image and video generation tasks. This innovative approach enhances the modeling of motion sequences by considering both past and future context during token prediction. Specifically, we predict all masked tokens simultaneously, retaining those with high confidence while re-masking others for re-prediction.  The non-autoregressive nature of this approach allows orders-of-magnitude faster sampling, generating a motion typically in 12-24 steps per codebook as opposed to hundreds of steps in autoregressive transformers and diffusion models. To facilitate long-form generation, we initialize new generations with tokens from the previous iteration. We integrate spatial attention in each transformer layer to reinforce the spatial relationship between multiple codebooks, thereby enhancing cohesiveness. To further ensure consistency and reduce unnatural motion, we introduce a token critic\cite{tokencritic,selfcritic} mechanism that guides the sampling process along with a text-motion alignment model for enforcing text consistency. We model the local motion and global translation separately and introduce three avenues for conditioning: cross-attention, input interpolation, and Feature-wise Linear Modulation (FiLM\cite{film}) layers. These conditioning mechanisms offer versatility and can be applied to various forms of conditions, including video or motion. 

Finally, we also aim to address fundamental questions overlooked by prior research. For instance, we investigate whether improving the tokenizer, as demonstrated in \cite{magvit2, stablediff}, results in significant quality improvements in motion generation. CLIP\cite{clip} has been the de-facto text encoder for motion generation even though it operates in the image-text latent space. Thus, we explore alternative representations for text embeddings, such as pooled\cite{bert,clip} versus full representations\cite{T5}, and the potential benefits of leveraging large language models like T5\cite{T5}. Additionally, we examine the representation of audio, considering whether deep learning-based methods generalize better to in the wild examples and speech compared to traditional spectrogram based approaches.
In summary, our contributions are:
\begin{itemize}
  \item Introduction of a novel motion generation framework capable of producing whole-body motion sequences of arbitrary length, conditioned on multiple modalities simultaneously, and adaptable over time.
  \item Utilization of three Vector Quantized Variational Autoencoders dedicated to modeling the local motion representation of body and hands separately, enhancing the granularity of motion representation. Predicting global root translation from local motion parameters. 
  \item Implementation of a bidirectional Masked Language Modeling (MLM) strategy, enabling parallel decoding of all codebooks simultaneously, thus improving processing efficiency. Integration of spatial attention mechanisms, a token critic, and a text-motion alignment model to ensure coherence and consistency in the generated motions.
  \item Conditioning of motion generation on both text and audio inputs. 

\end{itemize}

We evaluate our models on both full motion and conditional ablations on body only motion using popular motion generation evaluation metrics.

\section{Related Work}
\subsection{Vector Quantization}
The Vector Quantized Variational Autoencoder(VQ-VAE)\cite{vqvae} as an extension to VAE\cite{vae} by learning a discrete latent space instead of a continuous normal distribution. VQ-VAE's have shown promising results in generative tasks across various domains, such as image synthesis \cite{hierarchical, vqgan, vqvae2,muse,maskgit} and video generation\cite{phenaki, magvit,magvit2}, while Residual Vector Quantization (RVQ)\cite{soundstream}, a varient of VQ-VAE is used in audio compression and generation\cite{jukebox, audiolm, musiclm, musicgen, audiogen, soundstream,encodec}. VQ-VAE's have been also used to model motion\cite{t2mgpt, bailando,tomato,emage,momask} successfully.

\subsection{Motion Synthesis}

\paragraph{Text conditioned motion generation}Early approaches focused on learning a joint motion-text representation through transformer-based VAE\cite{actor,temos,humanml3d} or contrastive approaches\cite{tmr,motionclip, ohmg} that generate novel motion by sampling from a shared latent space. Modern text-based models leverage diffusion principles\cite{mdm,mld,motiondiffuse} or language model-based methods\cite{tomato,t2mgpt,momask,tm2t,motiongpt}.Language models typically adopt a two-stage approach: encoding motion data into a discrete space and subsequently employing an autoregressive or bidirectional transformer model to generate motion indices. These models are often conditioned on CLIP text embeddings. MoMask\cite{momask} use a RVQ\cite{soundstream} with multiple codebooks, where the first codebook is predicted using a masked language model(MLM), while subsequent codebooks are predicted using an autoregressive transformer. NeMF\cite{nemf} uses a continuous motion field represented by a VAE architecture.  Whole body motion generation combining body and hands is still in its nascent stage; HumanTOMATO\cite{tomato} employs a hierarchical vector quantized variational autoencoder (\(H^2VQ)\) with multiple codebooks and utilizes an autoregressive transformer to predict a flattened codebook sequence. They leverage text embeddings from Text-Motion-Retreival model(TMR)\cite{tmr} to enforce motion-text alignment. Notable datasets in this domain include HumanML3D\cite{humanml3d} for body-only generation, and the MotionX\cite{motionx} dataset, which further extends this in the SMPLX\cite{smplx} format for whole-body generation.
\paragraph{Music conditioned dance generation}Various network architectures have been proposed for music-driven motion generation, spanning 
CNN\cite{dancing2music,motionet,charmotsyn,learning2dance_gcn,bailando,music2dance}, RNNs/LSTMS\cite{groovenet,tempguidemusic,dancemelody,rhythmisadancer}, GANs\cite{dancing2music,deepdance}, reinforcement learning\cite{bailando}, motion graphs\cite{choreomaster}, diffusion models\cite{edge} and language models\cite{genrecond,aichoreo,youneverstop,attenspatiotemporal,dancerevolution, tm2d}. However, many of these approaches need specialised pre-processing to work with in-the-wild music\cite{bailando, edge}, require a seed motion\cite{aichoreo}, or have complex architectures\cite{choreomaster, dancing2music,danceformer}. EDGE\cite{edge} modifies diffusion based text-to-motion generation\cite{mdm} by cross-attending to jukebox\cite{jukebox} music embeddings. Common datasets include the AIST++\cite{aichoreo} dataset. 
\paragraph{Co-speech gesture generation}This involves generating full-body human gestures from speech audio and transcripts. The recent BEAT\cite{beat} dataset has enabled methods like \cite{emage,gesturediffuclip} to adapt text based motion generation architectures to this task.\cite{gesturediffuclip} adopts a diffusion framework to generate motion conditioned on text transcripts, audio, and optional style embeddings. It employs a learned gesture-text alignment model for embedding text, akin to the approach used in HumanTomato\cite{tomato}, which utilizes TMR\cite{tmr}. Audio features(onset and amplitude), are concatenated to the input, while text is integrated through cross-attention and style through AdaIN\cite{stylegan} layers. EMAGE\cite{emage} utilizes separate codebooks for the lower and upper body and the hands and uses a BERT\cite{bert} style model to reconstruct masked input motion instead of generating from scratch. While gesture generation models require a one-to-one correspondence between audio and text, our approach can generate motion conditioned on unrelated text and audio.
\subsection{Masked modelling for generation}

Masked Language Modeling (MLM), pioneered by BERT\cite{bert}, improves language understanding by training models to reconstruct masked inputs. Building upon this, MaskGIT\cite{maskgit} extended MLM to image generation tasks by introducing a variable masking rate during training and iteratively predicting tokens from fully masked inputs during inference. Muse\cite{muse} scales MaskGIT to 3B parameters and integrates it with the T5\cite{T5} language model for improved performance. MAGVIT\cite{magvit} enhances MaskGIT by introducing a 3D CNN-based VQGAN\cite{vqgan} tokenizer for spatial-temporal tokenization, while Phenaki\cite{phenaki} utilizes a ViViT\cite{vivit}-based tokenizer alongside Muse. MAGVIT2\cite{magvit2} further improves upon MAGVIT\cite{magvit} by enabling the learning of an exponentially larger codebook size. Addressing challenges in non-autoregressive generative transformers, Token Critic\cite{tokencritic} guides sampling by distinguishing between original and generated tokens. Token-Critic is used to select which tokens to accept and which to reject and resample. Self Token Critic\cite{selfcritic} proposes the addition of a binary prediction head into the model itself, allowing the model to evaluate the quality of generated tokens internally, thereby improving token generation quality.
Exploring masked language modeling (MLM) with multiple codebooks remains relatively unexplored. Existing MLM-based motion generation models, such as those in \cite{tomato, emage} typically operate on a flattened representation of the codebooks. MusicGen\cite{musicgen}, an auto-regressive transformer for music generation, extensively studies optimal codebook interleaving patterns. However, direct application of these findings in MLM-based generation is not straightforward due to the randomized un-masking scheme compared to unidirection auto regressive generation.




\section{System overview}

\paragraph{Pose Representation:} We use the representation specified in \cite{holden} used in Humanml3d\cite{humanml3d} and Motion-X\cite{motionx} datasets. We use the whole body Motion-X dataset and combine it with the BEAT\cite{beat} gesture dataset and Choreomaster\cite{choreomaster} dance dataset. We use the SMPLX\cite{smplx} representation with 52 joints (22 body and 30 finger joints) where the \(i\)-th pose is defined by a tuple of root angular velocity \(r^a\) along the Y-axis, root linear velocities \(r^x , r^y\) on XZ-plane, root height \(r^y\), local joints positions \(j^p\), velocities \(j^v\) and binary foot contact labels \(c\). The whole-body motion is represented as \(m_i = \{ r^a, r^x, r^z , r^y , j^p, j^v, c\} \in \mathbb{R}^{d_m} \) at 30FPS. We perform ablation studies on the inclusion of joint rotations.

\paragraph{Conditioning Representation:} We use the Encodec\cite{encodec} embeddings resampled to 30HZ as the audio conditioning signal and the T5-Large\cite{T5} LLM to extract text embeddings. We perform ablation studies with AIST++\cite{aichoreo} MFCC audio features, CLIP\cite{clip} text embeddings and CLAP\cite{clap} joint text-audio embeddings.

\begin{figure}
  \centering
  \includegraphics[width = \linewidth]{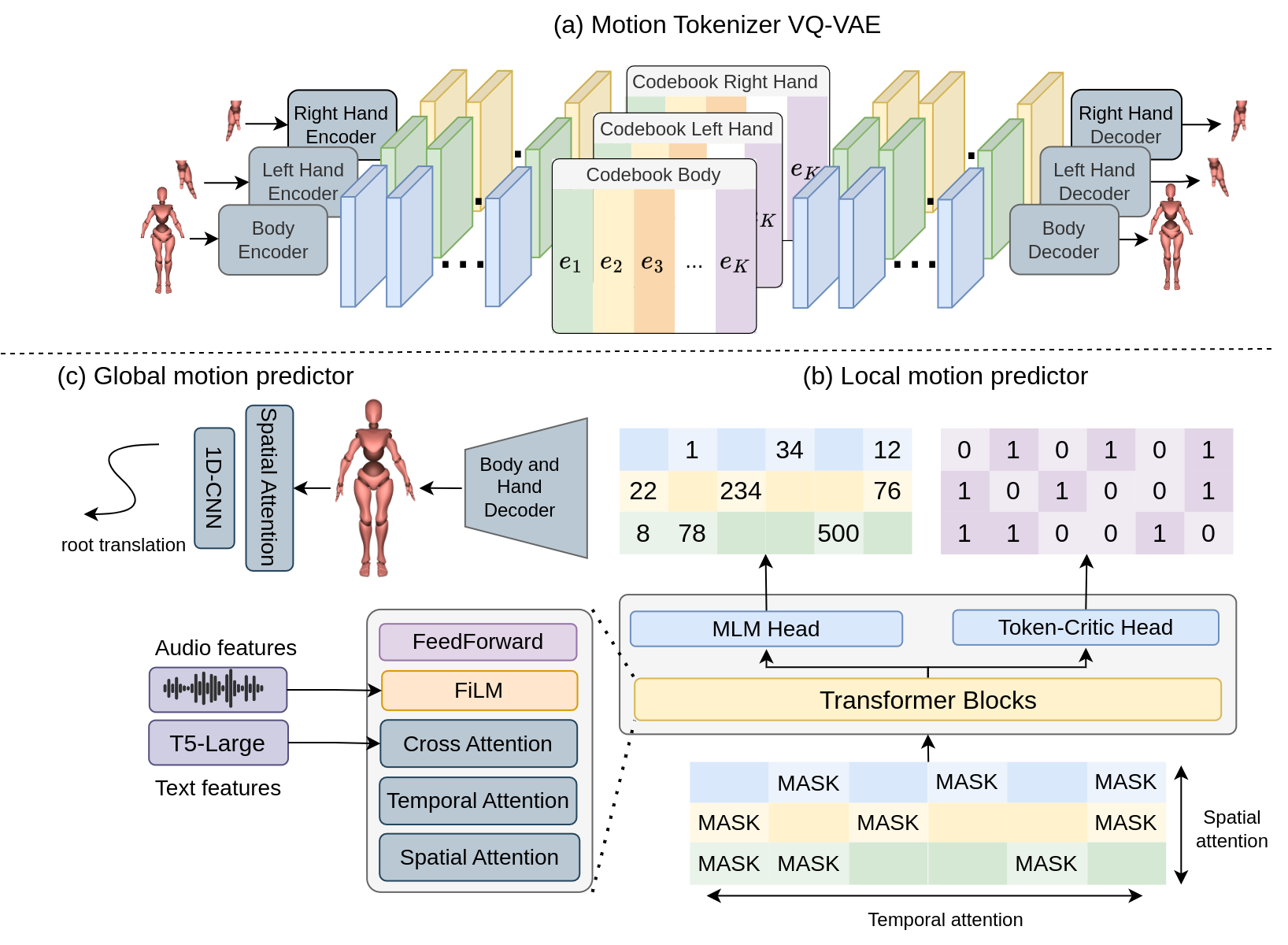}
  \caption{\textbf{Our 3 stage motion generation pipeline:} (a) Initial tokenization of the whole-body motion sequence, excluding translation, into three distinct motion sequences using VQ-VAEs dedicated to the body, left hand, and right hand. (b) During training, a random subset of tokens is masked in the input, and the model is tasked with predicting these missing tokens. A token critic is trained to discern between ground truth and predicted tokens. During inference, all motion indices in a sequence are simultaneously predicted, with the token critic guiding the decision on which indices to retain, remask, and resample. These indices are then mapped to the corresponding local motion using the VQVAE decoder. (c) A global motion predictor is trained to map body joint positions and velocities to root translation. During inference, this predictor is utilized to derive root translation from the predicted local motion.}
  \label{fig:full_pipeline}
\end{figure}
\subsection{VQVAE}
The VQ-VAE\cite{vqvae, vqgan} aims to learn a codebook \(C\) consisting of embeddings \(\{e_k \in \mathbb{R}^{d_c}\}^K_{k=1} \), where K is the number of codes of dimension \(d_c\) such that a motion sequence with \(L\) frames, \(X = [x_1,x_2,....,x_L] \) with \(x_i \in \mathbb{R}^{d_m}\), can be reconstructed back after passing through the autoencoder architecture and discretized by the codebook as shown in Figure \ref{fig:full_pipeline}. Passing the motion sequence \(X\) through the Encoder \(E\) results in latent features \(Z^e = E(X)\), with \(Z^e = [z^e_1,z^e_2,....,z^e_{l}] \), \(z^e \in \mathbb{R}^{d_c}\) with \(l/L\) being the downsampling ratio. 
\paragraph{Training objective:} For i-th latent feature \(z^e_i\), the quantization through the codebook is to find the most similar element in \(C\), which can be written as:

\begin{equation}
z^q_i = \underset{c_k \in C}{argmin} ||z^e_i - e_k ||_2
\end{equation}

A decoder \(D(Z^q)\) then decodes the embedding vectors back into the input space. The original formulation of the optimization goal\cite{vqvae} is:

\begin{equation}
\mathcal{L}_{vq} = \underbrace{L_{huber}(X , D(Z^q))}_{reconstruction}  + \underbrace{|| sg[Z^e] - Z^q||_2}_{codebook} + \underbrace{\beta||Z^e - sg[Z^q]||_2}_{commit}
\end{equation}

Where \(sg\) stands for the stop-gradient operator that has zero partial derivatives during back-propogation. The commit loss prevents the encoder output from growing arbitrarily by constraining the encoder to the codebook embedding space. \(L_{huber}\) corresponds to the huber loss between the input and reconstructed motion sequences. 

We adopt methods from \cite{vqvae2,vqgan,jukebox,soundstream} for replacing stale codes and k-means initialization. For enhanced codebook utilization during inference, we employ techniques from \cite{vqvae_affine} involving affine reparameterization of the codebook with a shared global mean and standard deviation and alternate optimization on the commit and reconstruction loss.

A single-codebook VQ-VAE yielded suboptimal results, and simply enlarging the codebook size poses computational and performance challenges\cite{soundstream,encodec}. While RVQ\cite{soundstream} offers a solution using multiple codebooks to iteratively reduce reconstruction error, we believe a more effective approach is to utilize separate codebooks for the body and both hands. It forces the motion generator to discern the relationship between body and hands, paving the way to further partition the body representation into finer segments. We use 3 VQ-VAE's corresponding to body, left hand, and right hand motion.


\subsection{Global Motion Predictor}
Inspired by \cite{nemf, emage}, we adopt a strategy to predict the global translation parameters conditioned solely on local motion parameters. However, we refine this approach by predicting the root XZ linear velocity separately while predicting the root orientation and height alongside the remaining motion parameters. Our rationale lies in the strong correlation between orientation, height, and the conditioning inputs. While \cite{nemf} employ skeletal convolutional layers \cite{skel_conv} to enforce spatial relationships with the nearest joint, our approach acknowledges the inherent coordination between limbs during activities like walking, where arms naturally synchronize with legs. To capture this coordination, we leverage spatial attention layers, allowing the model to learn the appropriate relationships between joints with higher flexibility and resolution.

\subsection{Local Motion Generator}

We encode motion of length \(L\) using the previously defined VQ-VAEs for the body, left hand, and right hand, each with a codebook of size \(K\), resulting in downsampled motion tokens \({m}_{(1:l)\times3}\) with shape \((l \times 3) \). Unlike previous approaches that flatten this sequence to \((3 \cdot l \times 1)\), we preserve the individual codebook tokens for each body part, maintaining them stacked. Prior to the attention layers, we augment each token sequence with positional sinusoidal embeddings and audio embeddings. The audio embeddings undergo preprocessing via a TCN layer. Spatial attention is then computed across the codebook dimension, while temporal self-attention operates across the sequence length. Conditioning is repeated three times, enabling distinct conditions for each body part during inference. Text conditions inform cross attention to ensure each motion embedding contains relevant textual information, while audio conditions are processed through FiLM\cite{film}layers to affine transform the motion sequence, aligning it with the audio input.

Next, we model the conditional motion token distribution using Masked Language Modeling (MLM) following prior work\cite{maskgit,muse,phenaki,momask,magvit}. Given a motion sequence \({m}_{(1:l)\times3}\), we employ a cosine scheduler to randomly mask \(l \cdot cos( \pi\tau_i/2 )\) tokens of each codebook with a special \([MASK]\) token at training step $i$ creating the masked motion sequence \(\bar{m}_{(1:l)\times3}\). \(\tau_i \in [0,1]\) is uniformly randomly sampled. Subsequently, we refine the model parameters \(\theta\) by  minimize the negative log-likelihood concerning these masked tokens, leveraging the encoded text (\(T\)), audio (\(A\)) embeddings, and unmasked tokens: 


\[ L_{\text{MLM}} =  - \sum_{j=1}^3 \sum_{\substack{i=1 \\ \bar{m}_{i} = [MASK]}
}^l  \log p_{\theta}(m_{i,j} | \bar{m}_{(1:l)\times3}, T , A) \]

where the negative log-likelihood is computed as the cross-entropy between the ground-truth
one-hot token and predicted token. The \([MASK]\) tokens in  \(\bar{m}_{(1:l)\times3}\) are then replaced by the predictions of the motion generator to give \(\hat{m}_{(1:l)\times3}\). The token-critic parameterised by \(\phi\)  discern between configurations of tokens likely belonging to the real distribution \(y_{(1:l)\times3}\) and those generated by the model by optimising the binary cross entropy loss (BCE):

\[ L_{\text{TokenCritic}} =  \sum_{j=1}^3 \sum_{i=1}^l BCE \left( y_{i,j}, {\phi}(\hat{m}_{i,j} |  T , A) \right) \]
The token critic shares all weights with the motion generator except the last, where it uses a binary prediction head. We also incorporate classifier-free guidance (CFG)\cite{classifierfree,muse,phenaki} during training by randomly dropping the text and audio condition 20\% of the time.

\subsection{Inference}

We initialize all motion tokens as \([MASK]\). During each inference step, we simultaneously predict all masked motion tokens, conditioned on various combinations of text and audio embeddings along with previously predicted motion tokens. The scores predicted by Token-Critic are used to select which token predictions are kept, and which are masked and resampled in the next iteration. We compute conditional logits \( c \) and unconditional logits \( u \) for each masked token. The final logits \( g \) are derived by adjusting the unconditional logits by a factor of \( s \), known as the guidance scale:
\[ g = (1 + s) \cdot c - s \cdot u \]
After decoding the indices though the VQ-VAE's we predict the motion translation using the global motion predictor. 

\subsection{Implementation details}

Our implementation, based on PyTorch\cite{pytorch}, is trained on a Nvidia A40-48GB GPU, with inference on a Nvidia 2080ti. We use a batch size of 400 for the VQVAE and 200 for the TMR and motion generator. The VQ-VAE's encoder and decoder employ 1D TCNs with depth 8, dimension 768, and a codebook with 512 codes, downsampling factor of 4. Body VQ-VAE codes are of dimension 512, while hand codes are 256. The motion MLM model includes 8 transformer blocks with dimension 512, condition dropout 0.4, and FiLM layers every third block trained on a sequence length of 30 tokens. The VQ-VAE and motion generator have 176M and 45M parameters, respectively. We use Adam optimizer with LR 3e-4, cosine decay, and linear warmup. During inference, CFG scale is 6, sampling temperature is 0.4, with 24 iterations with overlap 10 frames during long duration generation.

\section{Expermiments}

\subsection{Quantitative results}

We assess the generated motions quantitatively from three perspectives introduced in \cite{t2mgpt}. Firstly, we evaluate the quality of the generated motions by measuring the Frechet Inception Distance (FID), which quantifies the disparity between the distributions of the generated and real motions. Secondly, we examine the alignment between texts and generated motions using the Matching-score to gauge the similarity between texts and generated motions, and R-Precision to determine the accuracy of motion-to-text retrieval within a pairwise motion-text set of size 32. Thirdly, we assess generation diversity by calculating the Diversity metric, which measures the average Euclidean distances among 300 randomly sampled motion pairs, and the MModality metric, which evaluates the diversity of generation within the same given text. Following a methodology similar to \cite{gesturediffuclip,tomato}, we train a text-motion retrieval model using TMR \cite{tmr}, modified to employ T5-Large \cite{T5} as both the text encoder and the sentence encoder \cite{sentenceT5}. We display the results in Table \ref{table:gen_results}. GPVC/GPRVC corresponds to whether the motion has rotations or not, base has 45M parameters while large has 125M parameters. We experiment with 72 steps and 3 steps.

\begin{table}
  \caption{Text to motion generation results on HumanML3D\cite{humanml3d} test set. Batch size 32. }
  \label{table:gen_results}
  \centering
    \begin{adjustbox}{width=1\textwidth}
  \begin{tabular}{lllllll}
    \toprule
    Model &FID $\downarrow$ & \multicolumn{3}{c}{R-Precision}$\uparrow$   &MM-Dist $\downarrow$   & Diversity $\rightarrow$       \\
    \cmidrule(r){3-5}
        & & Top-1 &Top-2 &Top-3   & &  \\
    \midrule
    \noalign{\vskip-\aboverulesep}
    Real motion & 0.0  & 0.290 & 0.5725 & 0.836 &  0.6636 &   1.3745    \\
    GPVC-72(base) & 0.1728 & 0.331 & 0.5325 & 0.661 & 0.9351 &  1.3326    \\
    GPVC-3(base) & 0.6121 & 0.0518 & 0.095 & 0.133 & 1.3726 &  1.3172    \\
    \bottomrule
  \end{tabular}
    \end{adjustbox}
\end{table}








\subsection{Qualitative results}

\begin{figure}
  \centering
  \includegraphics[width = \linewidth]{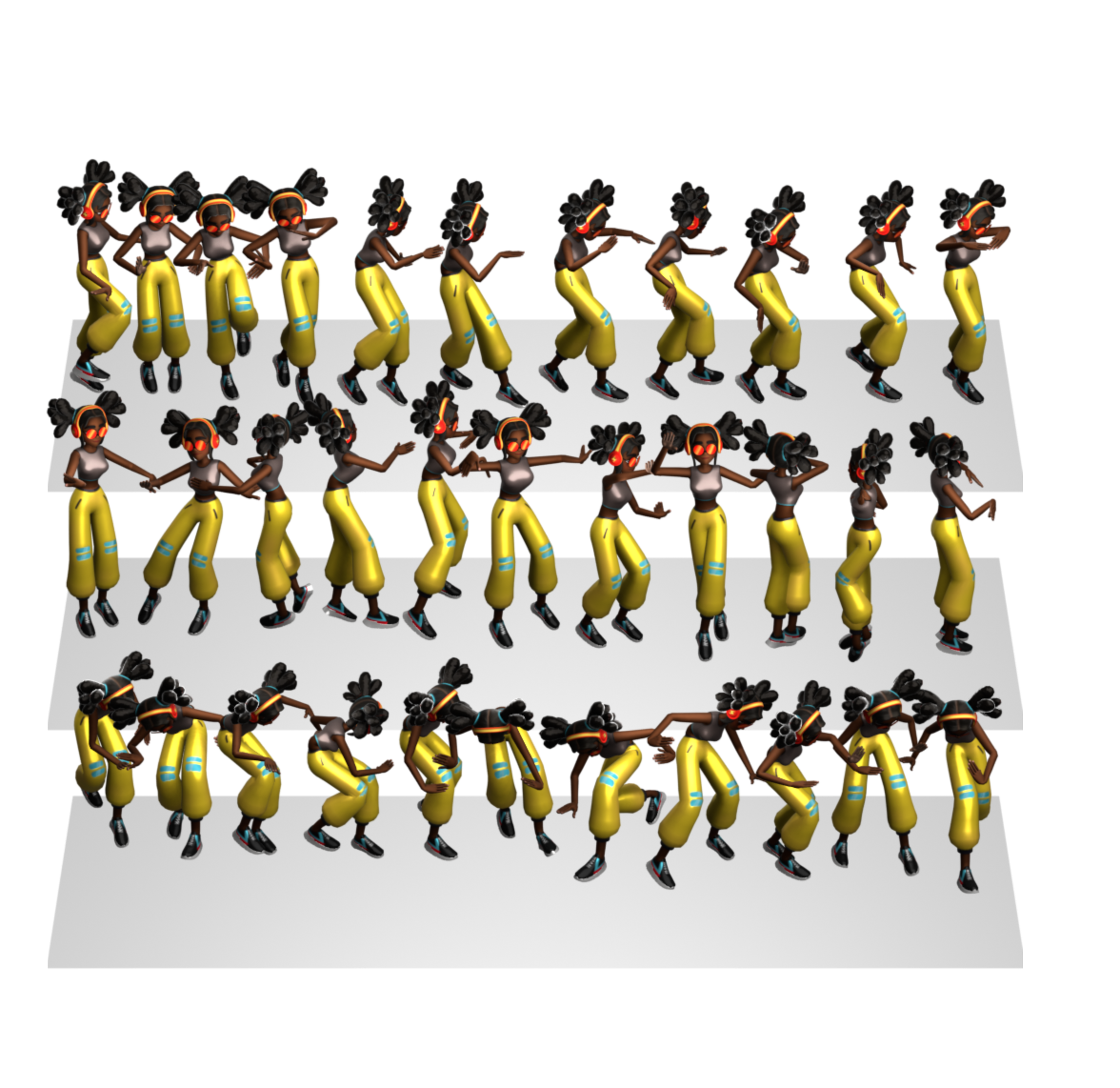}
  \caption{\textbf{Visual results on audio and text conditions:}  From top to bottom: Dance generated on break dance music, Dance generated on break dance music along with the text "a person doing ballet", The text "a person sneaks away while walking sideways". Only key frames are shown.}
  \label{fig:qual_results}
\end{figure}

We show qualitative results of our model on audio, text and audio + text conditioning in Figure \ref{fig:qual_results}. In the second row, we can see that the model faithfully follows both audio (break dance music) and text ("do a ballet") by introducing ballet turns intermittently. 

\subsection{Ablations}

We compare different VQ-VAE configurations in Table \ref{table:vqvae_results}, GPVC/GPRVC corresponds to whether the motion has rotations or not, base has 88M parameters while large has 225M parameters, A corresponds to using affine codebook training\cite{vqvae_affine}.
\begin{table}
  \caption{VQ-VAE motion reconstruction results on Motion-X\cite{motionx} test set. Batch size 256. }
  \label{table:vqvae_results}
  \centering
    \begin{adjustbox}{width=1\textwidth}
  \begin{tabular}{llllllll}
    \toprule
    Model &FID $\downarrow$ & \multicolumn{3}{c}{R-Precision}$\uparrow$   &MM-Dist $\downarrow$   & Diversity $\rightarrow$ Perplexity$\uparrow$                \\
    \cmidrule(r){3-5}
        & & Top-1 &Top-2 &Top-3   & & & \\
    \midrule
    \noalign{\vskip-\aboverulesep}
    Real motion & 0.0  & 0.2103 & 0.378 & 0.5397 &  0.6600 &  1.3918  & --  \\
    GPVC(base) & 0.0152 & 0.1779 & 0.332 & 0.4819 & 0.7368 & 1.397 & 265    \\
    GPVC-A(base) & 0.0144 & 0.1836 & 0.337 & 0.484 & 0.7278 & 1.3809 & 326    \\
    GPVC-A(large) & 0.0098 & 0.19 & 0.345 & 0.5 & 0.7368 & 1.3748 & 342    \\
    
    \bottomrule
  \end{tabular}
    \end{adjustbox}

\end{table}





\subsection{Applications}
Our motion generation framework can seamlessly stitches motion segments to create longer sequences and enables the generation of motion sequences of arbitrary length with consistent transitions. Demonstrating its versatility, we generate a 3-minute dance motion conditioned on YouTube music, showcasing its capability for long-form generation. Additionally, leveraging text prompts allows for directed motion generation, enabling smooth transitions between different actions. Furthermore, the bidirectional nature of our model supports motion completion or inpainting tasks, making it useful for motion editing and synthesis.

\section{Conclusion}

In conclusion, we have presented a novel motion generation framework that addresses key challenges in existing approaches. By adopting a bidirectional Masked Language Modeling (MLM) strategy, we achieve significant improvements in processing efficiency, enabling orders-of-magnitude faster sampling compared to autoregressive models. Our framework integrates spatial attention mechanisms and a token critic to enhance coherence and consistency in the generated motions. Moreover, we introduce separate modeling of local motion and global translation, along with versatile conditioning mechanisms, allowing for adaptation to various modalities and conditions. These contributions pave the way for the generation of whole-body motion sequences of arbitrary length, conditioned on multiple modalities simultaneously, and adaptable over time, opening up exciting possibilities for applications in fields such as animation, virtual reality, and human-computer interaction.

\bibliographystyle{plainnat}
\bibliography{neurips_2024.bib}

\end{document}